\documentclass[twocolumn,letterpaper]{IEEEAerospaceCLS}  

\usepackage[]{graphicx,float,latexsym,amssymb,amsfonts,amsmath,amstext,times}

\usepackage[]{graphicx}    
\usepackage{float}
\usepackage{multirow}
\usepackage{cite}
\usepackage{adjustbox}
\usepackage{ragged2e}
\usepackage{graphicx}
\usepackage{lipsum}
\usepackage[table,xcdraw]{xcolor}
\usepackage[hyphens]{url}
\newcommand{\ignore}[1]{}  

\begin{document}
\title{A Methodology to Assess the Human Factors Associated with Lunar Teleoperated Assembly Tasks}

\author{%
Arun Kumar\\ 
Center for Astrophysics and Space Astronomy\\
University of Colorado Boulder\\
Boulder, CO 80302\\
Arun.Kumar-1@colorado.edu\\
\\ 
Benjamin Mellinkoff\\
Center for Astrophysics and Space Astronomy\\
University of Colorado Boulder\\
Boulder, CO 80302\\
Benjamin.Mellinkoff@colorado.edu\\
\\
Wendy Bailey Martin\\
Engineering Management Program\\
University of Colorado Boulder\\
Boulder, CO 80302\\
Wendy.Bailey@colorado.edu
\and
Mason Bell\\
Center for Astrophysics and Space Astronomy\\
University of Colorado Boulder\\
Boulder, CO 80302\\
Mason.Bell@colorado.edu\\
\\ 
Alex Sandoval\\
Ball Aerospace\\
10 Longs Peak Drive\\
Broomfield, CO 80021\\
Asandov2@ball.com\\
\\
Jack Burns\\
Center for Astrophysics and Space Astronomy\\
University of Colorado Boulder\\
Boulder, CO 80302\\
Jack.Burns@colorado.edu
\thanks{\footnotesize 978-1-7281-2734-7/20/$\$31.00$ \copyright2020 IEEE}              
}

\maketitle

\thispagestyle{plain}
\pagestyle{plain}

\begin{abstract}
Low-latency telerobotics can enable more intricate surface tasks on extraterrestrial planetary bodies than has ever been previously attempted. In order for humanity to create a sustainable lunar presence, well-developed collaboration between humans and robots is necessary to perform complex tasks. This paper presents a methodology to assess the human factors, situational awareness (SA) and cognitive load (CL), associated with teleoperated assembly tasks. Currently, telerobotic assembly on an extraterrestrial body has never been attempted, and a valid methodology to assess the associated human factors has not been developed. The Telerobotics Laboratory at the University of Colorado-Boulder created the Telerobotic Simulation System (TSS) which enables remote operation of a rover and a robotic arm. The TSS was used in a laboratory experiment designed as an analog to a lunar mission. The operator's task was to assemble a radio interferometer. Each participant completed this task under two conditions, remote teleoperation (limited SA) and local operation (optimal SA). The goal of this experiment was to establish a methodology to accurately measure the operator's SA and CL while performing teleoperated assembly tasks.  A successful methodology would yield results showing greater SA and lower CL while operating locally. Performance metrics measured in this experiment showed greater SA and lower CL in the local environment, supported by a 27\% increase in the mean time to completion of the assembly task when operating remotely. Subjective measurements of SA and CL did not align with the performance metrics. This brought into question the validity of the subjective assessments used in this experiment when applied to telerobotic assembly tasks. Results from this experiment will guide future work attempting to accurately quantify the human factors associated with telerobotic assembly. Once an accurate methodology has been developed, we will be able to measure how new variables affect an operator's SA and CL in order to optimize the efficiency and effectiveness of telerobotic assembly tasks.


\end{abstract}
\newpage

\tableofcontents
\newpage

\section{Introduction}
    
    NASA has proposed the Artemis program that aims to put humans back on the Moon by 2024 for the first time since the Apollo program. Furthermore, NASA plans to create a sustainable human presence on the Moon by 2028 \cite{artemis}. Current technology is not advanced enough to allow astronauts to safely perform the majority of tasks necessary to create a sustainable lunar presence. However, collaboration between humans and robots can accomplish tasks that are too difficult or risky for astronauts. For example, there is evidence of water ice in shadowed craters near the lunar south pole \cite{ice}. This ice can provide great scientific insights about the Moon and serve as a valuable resource during lunar missions. However, the crater temperature varies between 40K and 110K which is outside the operating conditions of current space suits \cite{CraterTemp}. Robots could be sent into these craters to extract the ice without the need for direct astronaut involvement. Currently, the latency between the Earth and the Moon hinders the performance (time to completion) of telerobotic tasks \cite{Ben}. To fully leverage planetary robotics for lunar colonization, humanity needs infrastructure on the Moon and in lunar orbit.
    
    NASA's first step in creating a sustainable human lunar presence is the construction of a lunar and deep space research and exploration laboratory. Astronauts are limited to 21 days aboard the Orion crew capsule so to enable long duration lunar missions, NASA is currently constructing a space station to orbit the Moon, the Gateway. The Gateway will be launched to Near-Rectilinear Halo Orbit (NHRO) \cite{gerstenmaier}. In NHRO, the gateway will complete an orbit every seven days, on the seventh day drawing closest to the moon's surface. From NHRO, the Gateway can serve as a communication relay between Earth ground stations and the unexplored lunar farside.

    The presence of astronauts on the Gateway would enable low-latency teleoperation of lunar rovers. When the orbit of the Gateway is at the equivalent distance of Earth-Moon L2 approximately 60,000 km from the lunar surface, the expected latency between the Gateway and the lunar surface will be 0.4 seconds. This latency is within the human cognitive horizon, meaning operators on the Gateway will notice a slight delay when controlling surface assets, but the delay will not significantly hinder performance \cite{GatewayLatency}. The minimal delay between the Gateway and the lunar surface enables more advanced surface telerobotic tasks than has ever been attempted on an extraterrestrial body.
    
    Our research team is involved with a scientific mission requiring intricate surface telerobotics, FARSIDE (\textit{Farside Array for Radio Science Investigation of the Dark Ages and Exoplanets}). FARSIDE is a concept mission designed to place a low radio frequency interferometric array on the farside of the Moon \cite{Farside,FARSIDE2}. The lunar farside is the only location in the inner solar system free of human-generated radio frequency interference \cite{RFI}. A radio interferometer on the lunar farside could probe the Dark Ages and the Cosmic Dawn of the universe. The mission design requires a rover and a lander. The rover would be teleoperated to deploy antenna nodes from the lander on to the lunar surface.

    \subsection{Assembly Experiment}
    
    This paper details the methodology and execution of a small scale laboratory simulation of the FARSIDE mission. Since telerobotic assembly tasks have never been attempted on an extraterrestrial body, this research serves to determine a methodology to accurately assess the situational awareness (SA) and cognitive load (CL) of an operator while performing telerobotic assembly tasks analogous to a robotic lunar mission. Data from this experiment may serve as a baseline for future experiments in which hardware, software, and procedural changes will be employed in order to more accurately assess the human factors involved with telerobotic assembly. Once an accurate methodology has been developed, researchers can optimize human performance during telerobotic assembly operations by measuring how changes to a telerobotic system affect the associated human factors.
    
\section{Related Work}

    \subsection{Current State of Surface Telerobotics}
    
    Surface telerobotics on planetary bodies has been primarily used for reconnaissance and geological exploration tasks due to the high latency between the robot and the operator. Robotic precursor missions will be a necessary part of the Artemis program to determine how astronauts should allocate their limited time on the lunar surface. Robotic reconnaissance increases scientific understanding of the area, reduces travel risk, and improves astronauts' productivity and SA \cite{RoboticRecon,RoboticLunarRecon}. The Mars Exploration Rovers (MERs) utilized daily command sequencing to coordinate the rovers' tasks due to the high latency between Earth and Mars \cite{SpaceTelerobotics}. While command sequencing may not be the most efficient way to control a rover, the MERs demonstrated that command sequencing is an effective workaround to latency considering the rovers' longevity and scientific discoveries \cite{MERScience}. Since the Gateway will enable low-latency communication with the lunar surface, there is a unique opportunity to push the boundaries of lunar robotics by performing more intricate surface tasks than anything previously attempted. In present day, telerobotics is commonly used for in-space assembly tasks. The Special Purpose Dexterous Manipulator (SPDM) on the ISS consists of two symmetrical seven-joint arms that are teleoperated to perform in-space assembly tasks such as changing batteries, replacing cameras, and servicing satellites \cite{SpaceTelerobotics, SPDM}. In the coming decade, robotic assembly missions will play a major role in establishing the necessary infrastructure on the lunar surface for a sustainable human presence. Habitats will need to be built and deployed on the surface. 3D printing with lunar regolith will allow for parts to be built on the Moon, rather than sent to the Moon reducing launch costs. These parts could be assembled to form larger structures or tools \cite{BuildingHabitats}. The biological risks of operating on the lunar surface without the necessary infrastructure make robots the best suited to assemble the initial infrastructure before humans spend long periods of time on the Moon. Our laboratory is pursuing research regarding surface telerobotic assembly tasks in order to determine the operational requirements for future planetary telerobotic assembly missions.
        
    \subsection{Human Factors Associated with Telerobotics}
    
    Human factors play a crucial role in assessing the quality of telerobotic systems. The functionality of teleoperated robots is limited by the SA and CL of the operator. A single operator error could result in complete mission failure. For certain telerobotic applications, this could mean the loss of a large investment in a space mission or the loss of a patient's life on the operating table. In order to improve teleoperated systems, it is necessary to accurately quantify the associated human factors. This allows researchers to relatively measure changes in SA and CL. SA and CL are multifaceted concepts which makes direct measurement difficult. Endsley describes SA as, ``the perception of elements in the environment within a volume of time and space, the comprehension of their meaning, and the projection of their status in the near future" \cite{SA}. CL can be defined as the workload on an operator's cognitive system caused by performing a task \cite{CLDef1,CLDef2}. Researchers have found a negative correlation between SA and CL \cite{SA&CL}. Performance metrics (most commonly task duration) are often used to gain insight on an operator's SA and CL, but there is no defined relationship between performance and objective measurements of SA and CL \cite{performance}. On the other hand, subjective measurements can be designed to address the multifaceted nature of human factors. Previous research has shown a positive correlation between subjective perceptions of telepresence and task performance \cite{hypothesis}. Sheridan defines telepresence as sensing information about the teleoperator (robot being controlled) and task environment and communicating the information to the human operator \cite{Sheridan}. This definition implies a positive correlation between telepresence and an operator's SA. By substitution, subjective measurements of SA positively correlate with performance metrics of telerobotic tasks, and subjective measurements of CL negatively correlate with performance metrics of telerobotic tasks. Typically surveys are used to subjectively measure human factors, however the validity and reliability of surveys can vary based on the application \cite{c4i,SAmeasurement}. Previous research has not determined a methodology to accurately assess the human factors associated with telerobotic assembly tasks. The correlation between subjective measurements of human factors and performance serves as a way to validate the methodology presented in this paper to assess an operator's SA and CL while performing teleoperated assembly tasks.

    \begin{figure}[H]
        \centering
        \includegraphics[width=0.48\textwidth]{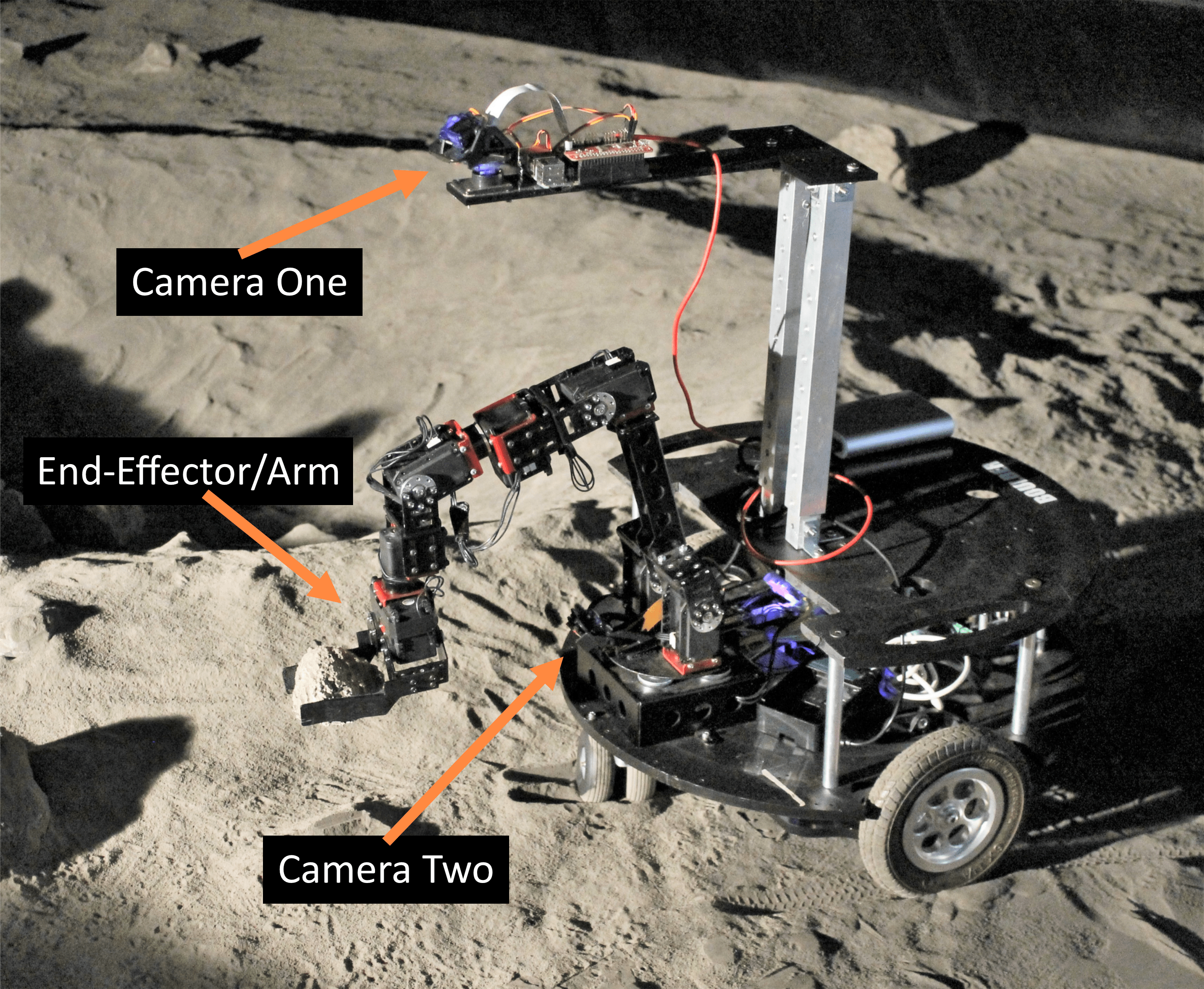}
        \caption{Armstrong Rover: Equipped with two Cameras and Crustcrawler Pro-Series Robotic Arm. This picture was taken at the NASA Ames Research Center in their SSERVI Regolith Testbed / Lunar Lab.}
        \label{fig:Armstrong}
    \end{figure}   
\section{Implementation}

    \subsection{Telerobotic Simulation System}
    
    The goal of this work was to define a way to assess an operator's SA and CL while performing teleoperated assembly tasks. To assess these factors, we required a mobile robotic system with the ability to pick and place objects as well as an interface that enables teleoperation of the robotic system. We named this robotic system and interface the Telerobotic Simulation System (TSS).
    
    For this experiment, a commercial-off-the-shelf rover was integrated with a 6 degree-of-freedom Crustcrawler Pro-Series Robotic Arm and two cameras to create the ``Armstrong Rover'' (Figure \ref{fig:Armstrong}). The rover's drive system is comprised of a two wheel differential drive with two casters for support. This provides the rover the ability to move forward, backward, and rotate in place. 

    The location of the cameras was an essential aspect of the TSS because it was the primary feedback from Armstrong to the operator. The TSS only utilized two cameras. These cameras are sufficient for assembly, as more than one view allows for the visualization of depth within the field of view \cite{cams}. Two video streams gave the operator enough feedback to mentally form a 3D landscape without overwhelming the operator with excessive feedback. Camera one was placed on a stand above the robotic arm with pan and tilt capabilities. This allowed the operator to view the surroundings of the rover for navigation and provided an aerial view of the end effector which was highly effective when picking up objects. Camera two was placed on the base of the robotic arm with only tilt functionality. Since this camera was attached to the robotic arm, the camera would pan as the robotic arm moved. This ensured that the end effector could always be viewed from camera two. This camera was integral for placing objects. When the robotic arm is holding an object, the view of the ground from camera one is obstructed by the object. Feedback from camera two provides the operator with sufficient information to precisely place objects. Views from these cameras while deploying an object are shown in Figure \ref{fig:CamGUI}.
    
\begin{figure}[H]
            \centering
            \includegraphics[width=0.48\textwidth]{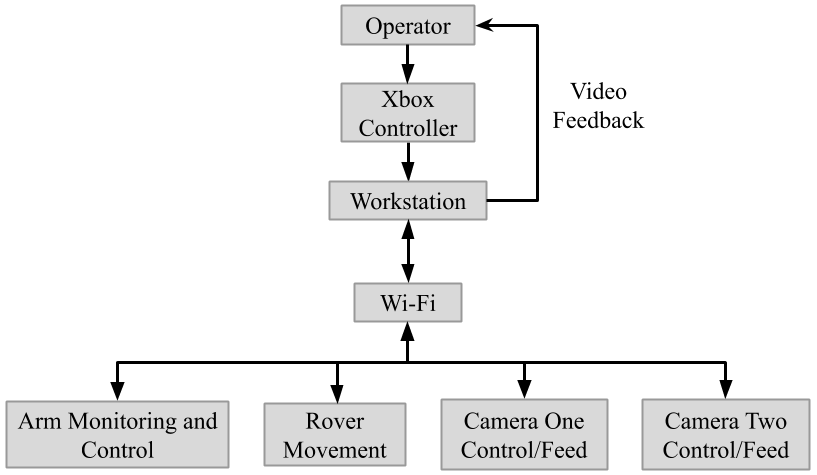}
            \caption{This system flow chart shows the connections between all the individual components of the TSS.}
            \label{fig:SBD}
        \end{figure}

\begin{figure}[H]
        \centering
        \includegraphics[width=0.47\textwidth]{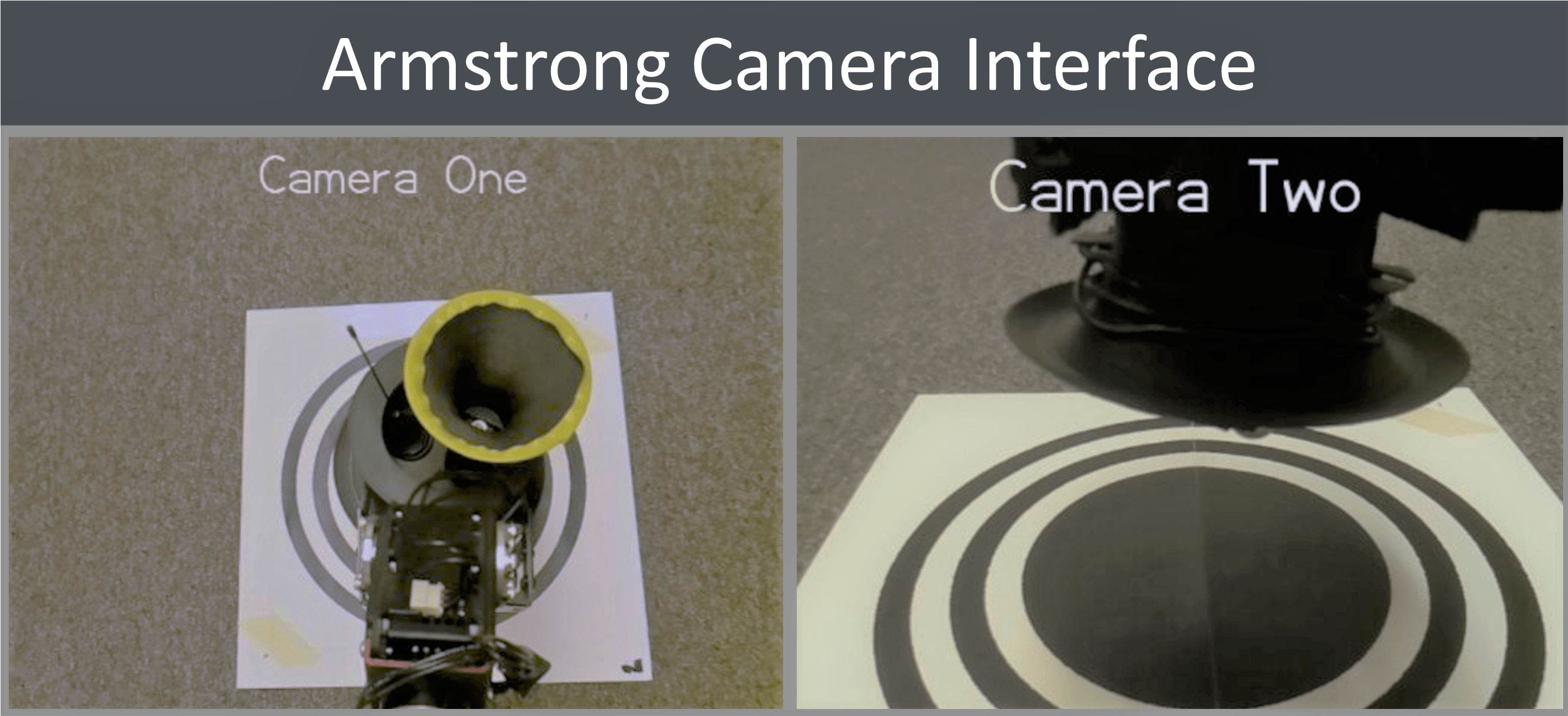}
        \caption{Dual view camera user interface providing the operator large real time imaging of the assembly in process} 
        \label{fig:CamGUI}
    \end{figure}

    Operators controlled Armstrong from a desktop computer using an Xbox controller. Inputs from the Xbox controller were sent from the desktop to a Raspberry Pi microcontroller on Armstrong via a local area network. Rover movement commands were serially transmitted from the Raspberry Pi to an Arduino which controlled the drive motors on Armstrong. Arm movement commands were processed on the operator's desktop using ROS and MoveIt. MoveIt solved the inverse kinematics necessary to allow the operator to move the end effector along the X, Y, and Z axes. Individual joint state values were sent from the desktop to the onboard Raspberry Pi and then applied to the individual joint actuators of the robotic arm. A visual of the data transfer throughout the TSS is shown in Figure \ref{fig:SBD}. 

\begin{figure}[H]
        \centering
        \includegraphics[width=0.4\textwidth]{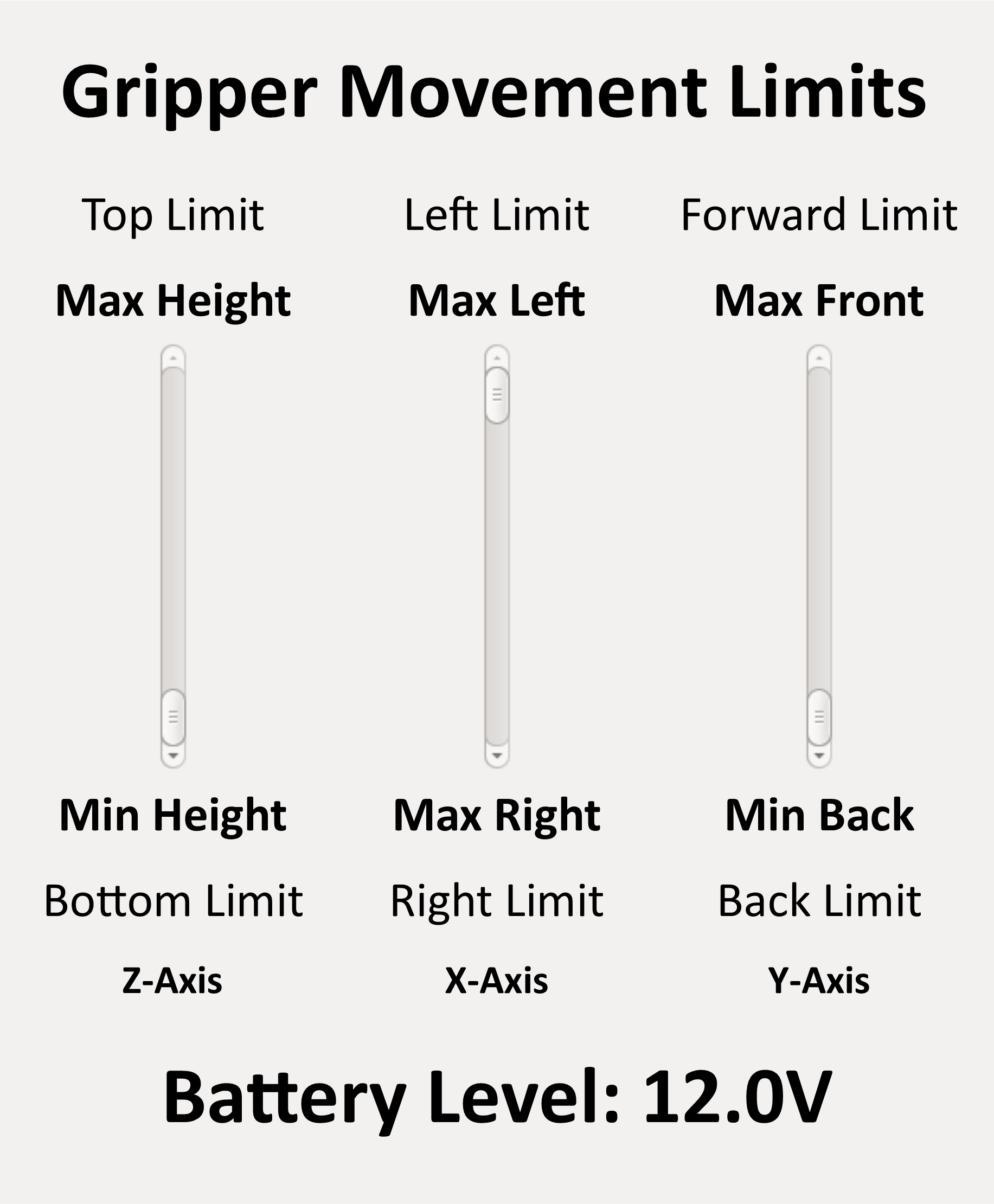}
        \caption{Gripper/Arm movement limits as part of the feedback user interface} 
        \label{fig:sliders}
    \end{figure}

\begin{figure}[H]
        \centering
        \includegraphics[width=0.4\textwidth]{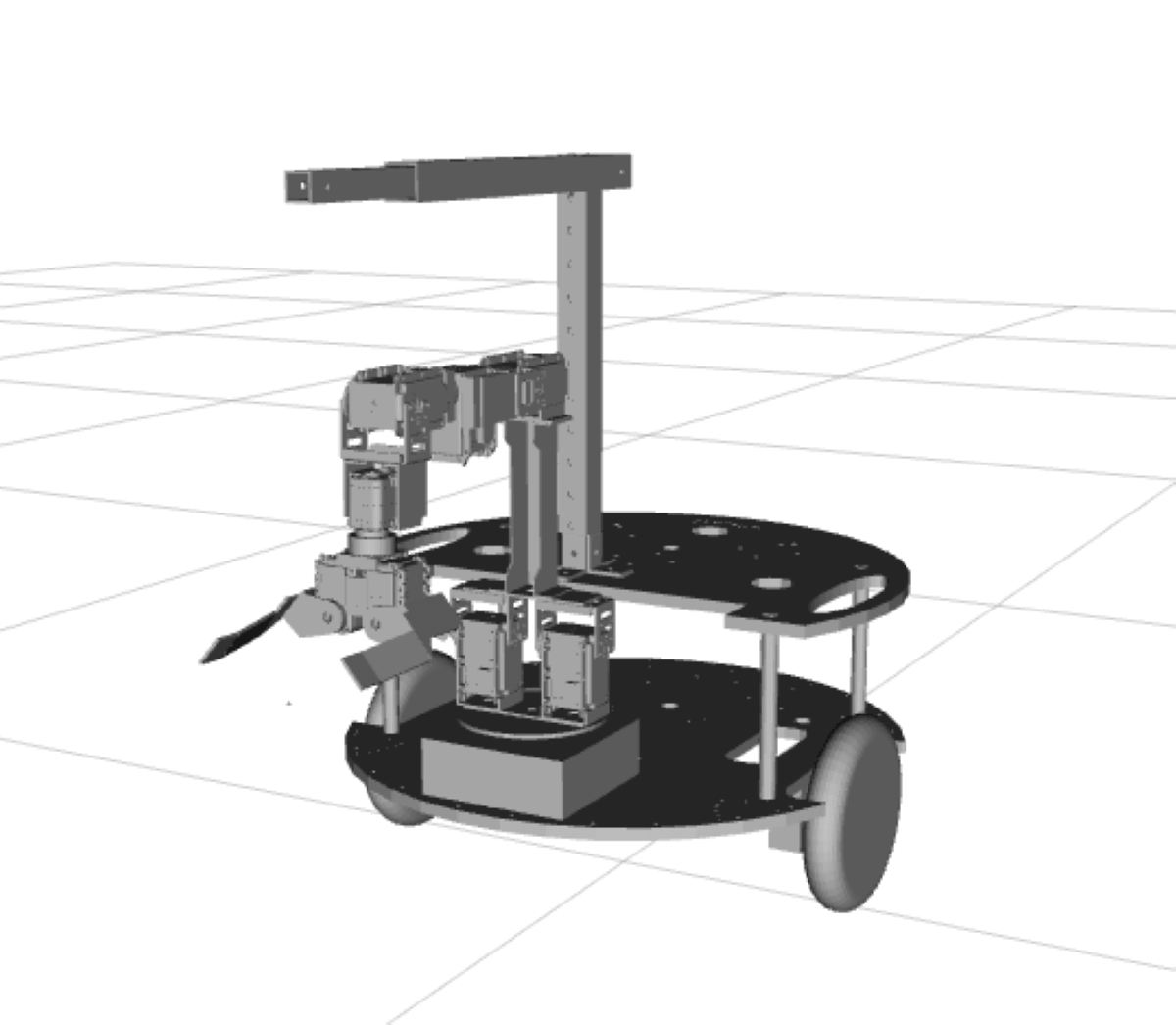}
        \caption{Real-Time updating rover model as part of the feedback user interface} 
        \label{fig:bot_gui}
    \end{figure}

    To prevent the operator from applying too much torque to the servo motors on the robotic arm, the end effector was confined to an invisible cube. The end effector's location with reference to the walls of the invisible cube was displayed on the user interface (Figure \ref{fig:sliders}). If the end effector reached one of the confining walls of the cube, the arm would stop moving and the Xbox controller would vibrate as feedback to the operator. Real-time joint state values of the robotic arm were sent back from the Raspberry Pi to the desktop and displayed on RViz, a ROS plugin that provides a 3D visualization of the current state of the robotic arm (Figure \ref{fig:bot_gui}). The interface also contained the video feeds from both cameras on the rover. The camera streams were displayed on a web page that allowed the operator to full-screen either camera or view both cameras simultaneously (figure \ref{fig:CamGUI}).

    \subsection{Radio Interferometer}
\begin{figure}[H]
    \centering
    \includegraphics[width=0.35\textwidth]{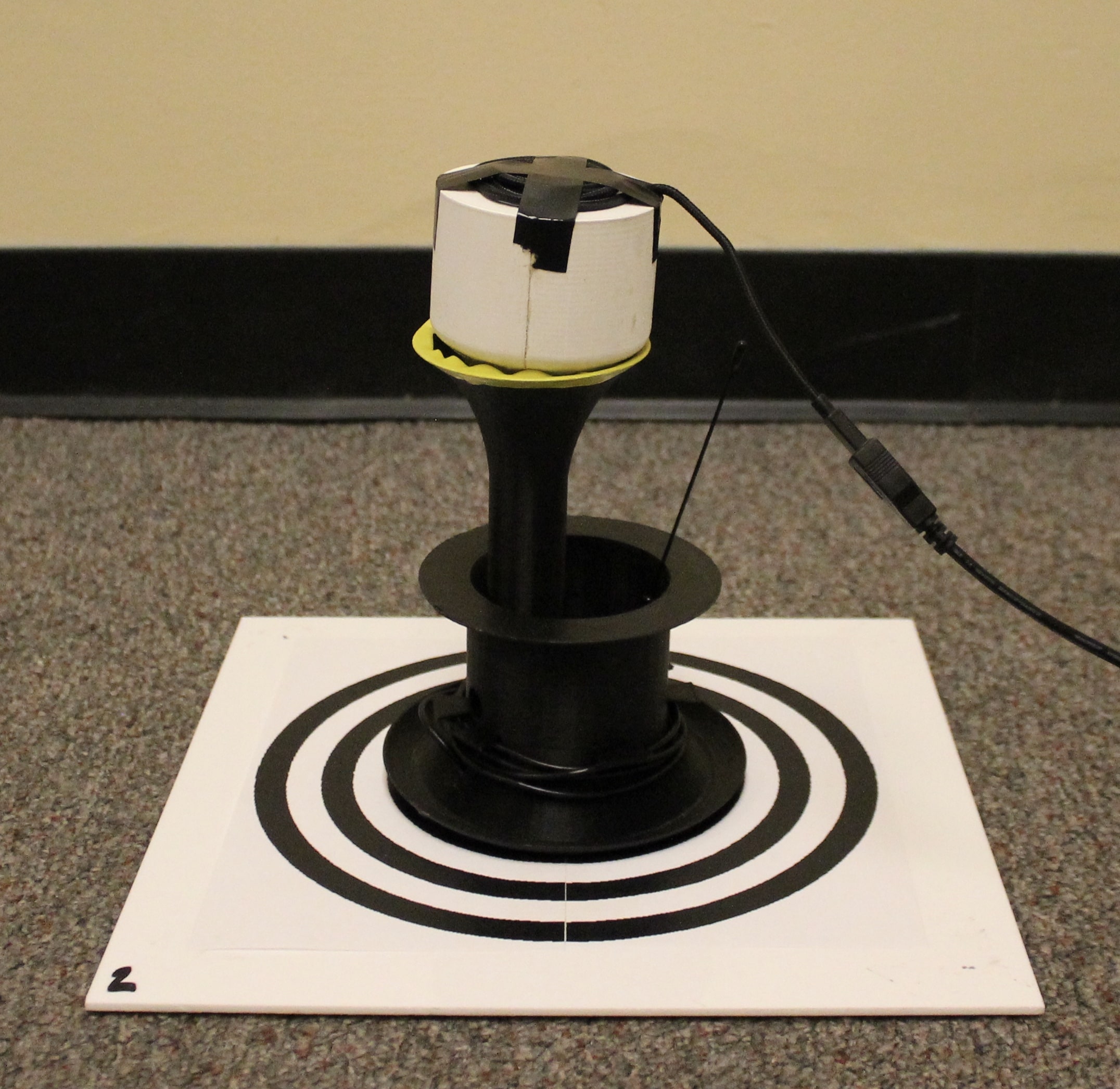}
    \caption{Fully assembled and deployed antenna unit. The black case is the antenna module. The white case is the USB module. The modules are connected via a magnetic micro USB.}
    \label{fig:AntennaUnit}
    \end{figure}

    The assembly task in this experiment was the construction of a radio interferometer. The radio interferometer used in this experiment consisted of three radio antennas. The three antennas are powered and transmit data via a magnetic micro USB connection. 3D printed modules were designed for the antennas and the micro USB cables (Figure \ref{fig:AntennaUnit}). The modules allow Armstrong to easily interact with the radio interferometer hardware. An antenna module houses one antenna and one half of a magnetic micro USB connection. A USB module houses the other half of the magnetic micro USB connection. Placing the USB module inside the trunk of the antenna module (as depicted in Figure \ref{fig:AntennaUnit}) creates an operational antenna unit. Three assembled antenna units create a radio interferometer. This assembly task requires the operator to precisely pick and place objects which is a large component of construction. However, this task does not encompass all facets of assembly.
    
\section{Experiment}

    In order to assess the human factors associated with telerobotic assembly tasks, we designed an experiment with human factors as the dependent variable and assembly location (remote teleoperation or local operation) as the independent variable. Participants completed the assembly task while remotely teleoperating Armstrong and locally operating Armstrong. Data collected during the experiment included performance metrics (specific metrics will be discussed later in this section) and subjective measurements of SA and CL.
    
    \begin{figure}[H]
        \centering
        \includegraphics[width=0.45\textwidth]{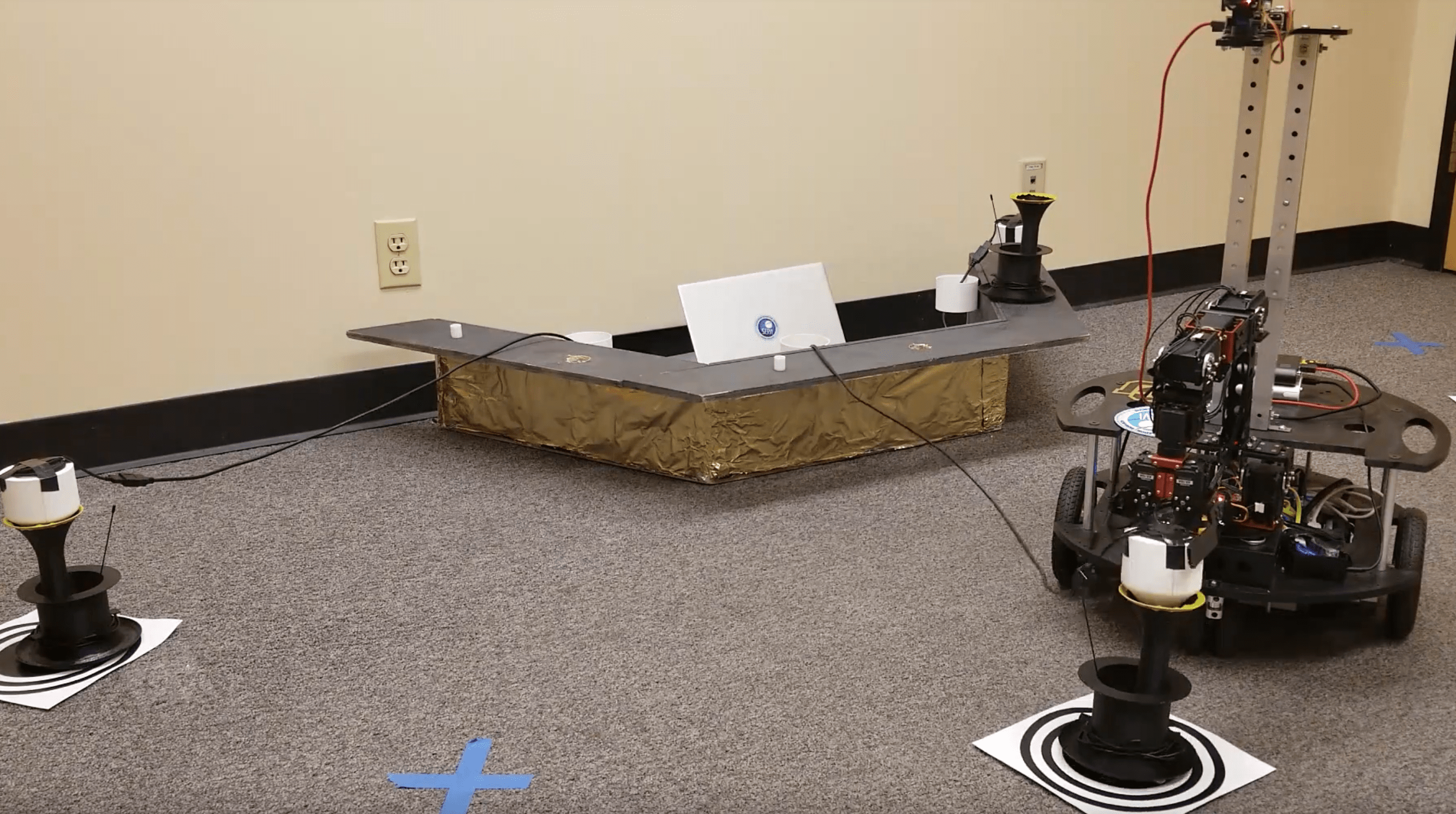}
        \caption{The Mock Lander equipped with the required parts for three antenna array units. The antenna modules are black, and the USB modules are white. Unit 1 has already been placed, unit 2 is being assembled, and unit 3 remains on the lander.}
        \label{fig:lander}
    \end{figure}

    \subsection{Setup}
    There were two separated rooms used in this experiment, the operations room and the command room. The operations room was where Armstrong executed the assembly task (Figure \ref{fig:lander}). In this room, a mock lander was placed against the wall with the necessary components to assemble three full antenna units. Three circular targets marked the deployment location for each unit. There was a computer station in operations room used for local operation of Armstrong. The command room also contained a computer station. This room was used for remote teleoperation of Armstrong. The participant's goal was to operate Armstrong to assemble a radio interferometer in both the local and remote environments.
    
    \subsection{Procedure}
    
    This experiment consisted of three phases: training phase, remote phase, and local phase. To begin, the participant was placed in the command room. The training session consisted of an explanation of the TSS, followed by a controls demo. First, the researcher transferred Armstrong from the operations room to the command room. All cameras on Armstrong were pointed out to the subject to give the subject an understanding of the video feedback system. Next, the researcher gave a verbal explanation of how to control the rover in addition to a handout with the button layout of the Xbox controller. The participant was allowed to refer to this handout at any point during the experiment. The operator was then given five minutes to practice the controls with the robot in sight. Next, we explained the assembly task to the participant. The participant was given an antenna module and USB module to obtain a clear understanding of each of the components and how they connect. The participant was instructed to attach the USB module to the antenna module with their own hands to ensure they understood the assembly task. With the rover now back in the operations room, we allotted 10 minutes for the subject to practice these controls while teleoperating Armstrong. Practice consisted of assembling one full antenna unit with no failure constraints or data gathering. After 10 minutes, the subject was instructed to stop entering commands.
    
    After the conclusion of the training phase, the participant moved on to either the remote phase or the local phase. Every odd numbered participant began in the remote phase, and every even numbered participant began in the local phase. During the remote phase, the subject was located in the command room teleoperating Armstrong in the operations room via the TSS. We instructed the participant to begin assembling and deploying the first antenna unit. The participant was allotted 15 minutes for the assembly and deployment of each antenna unit. We began timing as soon as the operator began inputting commands. There were 4 main sub-tasks that must be completed in the correct order for each antenna unit. (1) First, the antenna module must be picked up from the lander. (2) The antenna module must be deployed onto the specified target. We told the participant to aim for the center of the target. (3) Next, the USB case must be picked up from the lander. (4) Finally, the USB case must be mated with the antenna case via the magnetic micro USB connection. After mating the USB case and the antenna case, the antenna unit is complete. After assembling one antenna unit, the researcher moved Armstrong to the second starting location. The participant repeated this process two additional times in order to assemble the full radio interferometer. Two possible failure modes existed: if the operator dropped any of the components (antenna module or USB module) during assembly, or if the participant was unable to complete the assembly in the allotted 15 minutes. When the operator dropped a component, time was paused while the researcher returned Armstrong to the starting position and the dropped module to its starting location on the lander. We recorded the failure, and time resumed when the participant began to operate the rover again. If the operator exceeded the time limit, we recorded the failure, and the experiment proceeded as if the antenna installation was completed.
    
    During the local phase, the participant was brought into the operations room and placed at the computer station. The participant repeated the same assembly procedure as the remote phase, however they did not receive video feedback via the computer interface. Instead, we instructed participants to rely on their own vision while operating Armstrong. The same time limits and failure modes existed in the local phase as in the remote phase.
    
    \subsection{Performance Metrics}
    
    Performance metrics measured during the experiment included time to completion, number of failures, and antenna unit placement. Time to completion was recorded when the antenna module was deployed onto the target and when the micro USB connection was completed. Failures were logged by indicating the type of failure (dropped module or time limit exceeded), the assembly location (local or remote operation), the unit number (antenna unit one, two, or three), and the module type if a module was dropped (antenna module or USB module). The distance from the center of the antenna unit to the center of the target was measured to determine accuracy of deployment. We chose these metrics because they correlate to the operator's SA and CL \cite{hypothesis}.
    
    \subsection{Subjective Measurements}
    
    To more thoroughly assess the human factors associated with telerobotic assembly tasks, we employed subjective measurements of SA and CL in addition to the measured performance metrics. The Situational Awareness Rating Technique was used to measure SA, and the NASA Task Load Index was used to measure CL. Both surveys were administered electronically. A positive correlation between the performance metrics and the SART scores, and a negative correlation between the performance metrics and the TLX scores would serve as evidence that the methodology accurately quantifies human factors.

         \subsubsection{Situational Awareness Rating Technique}
         The Situational Awareness Rating Technique (SART) was administered twice. Once after the local phase and once after the remote phase. The SART is designed to measure the SA of a person during a task. This subjective rating is determined after the participant rates a series of questions on a scale of 1 to 7 (7 being the highest). The ten questions originate from three different domains: attentional demand, attentional supply and understanding \cite{SART}. The breakdown of these three categories into the ten respective questions is shown below in Table \ref{tab:sartchart}.
         
        \begin{table}[H]
            \centering
            \caption{This chart details the origins of the 10 question SART analysis. Each question stems from one of the three domains: Attentional Demand, Attentional Supply, Understanding}
            \fboxsep=0mm
            \fboxrule=0.25pt
            \fbox{\includegraphics[width=0.45\textwidth]{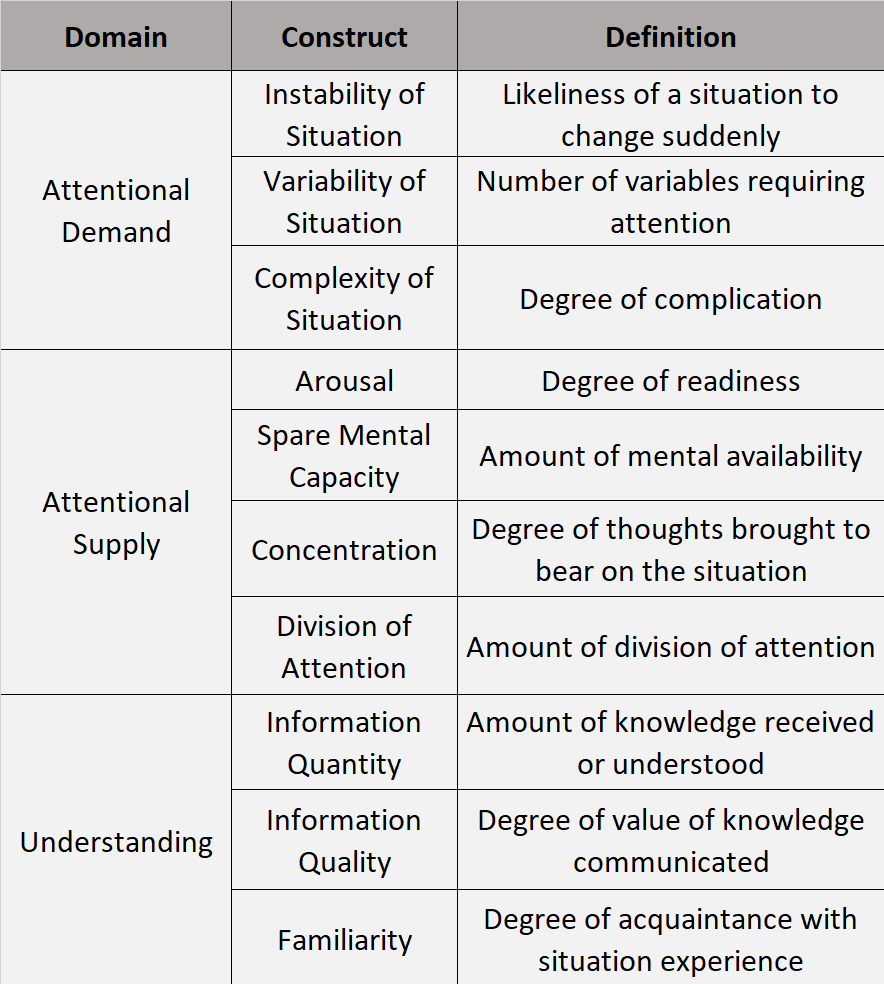}}
            \label{tab:sartchart}
        \end{table}
        
         After the participant has answered all ten questions, the researcher calculates the final SART score using the formula shown in equation \ref{eq:SART}, where U = Summed Understanding,  D = Summed Demand, and S = Summed Supply. Using the 
         outline in table \ref{tab:sartchart}, the summation for U, D and S are derived from the scores to their corresponding questions. A higher score correlates to higher SA \cite{SART}.

        \begin{equation} \label{eq:SART}
            SART = U - (D - S) 
        \end{equation}
         
        \subsubsection{NASA Task Load Index}
         
         The NASA Task Load Index survey was also administered after both the local and remote phase. We chose this survey because it is the most widely validated method of assessing cognitive workload. \cite{c4i} We used this survey to measure the participants overall workload during a task, by focusing on six different scales: Mental Demands, Physical Demands, Temporal Demands, Performance, Effort, and Frustration. There are two parts to each of the six scales, a weight and a rating. The weights are determined after asking the subject 15 pairwise comparisons of the 6 scales, covering each possible combination of the scales. Each comparison presented two of the scales, leaving the subject to chose which contributed more to the workload of the task. A tally sheet is used to keep track of the number of times the subject has picked each scale. After all of the comparisons, the number of tallies correlates to that scale's weight. The ratings are determined by having the user rate each scale on a range from 0 to 100. The higher the rating, the greater impact the subject felt that scale had on the task \cite{NASATLX}. Each scales rating can be multiplied by its weight to determine the weighted rating. To calculate the final task load index, the administrator follows the formula in equation \ref{eq:TLX}.
         \begin{equation}
            \label{eq:TLX}
             TLX = \sum \: Weighted \: Ratings \: \:  \div \: \: 15
         \end{equation}
         
         \subsubsection{System Usability Scale}
         
         The System Usability Scale survey is only administered after both the local and remote phases are completed. This scale rates the usability of the TSS. The ratings will be taken into account when considering areas of improvement for future iterations of this experiment. The SUS asks the participant to rate ten questions, related to ease of system use, on a scale of 0 to 5 (0 = Strongly Disagree, 5 = Strongly Agree). The final SUS score can be calculated by multiplying the sum of all ratings by two.

    \subsection{Demographics}
         
    15 participants between the age of 18-25 were recruited from The University of Colorado Boulder. All participants consented to the experiment, and the experiment was approved by the Institutional Review Board (IRB). Prior to beginning the experiment, each participant completed a demographic survey. Data collected details the participant's gender, age, robotics familiarity, and gaming experience.
    
    Based on empirical evidence from previous experimentation and the associated variation with the time-based evaluation of telerobotic tasks, the sample size necessary to detect a shift in the mean level of assembly time was calculated to be 15 participants. This calculation was based on two assumptions. First, the nature of the study was exploratory, meaning that the standard deviation related to assembly tasks prior to this work was not known. As such, it was assumed that the minimum treatment effect size would be roughly 1.5 times the estimate for the population standard deviation. Using this assumption, the power to detect a shift in the mean with a sample size of 15 was calculated to be 0.9531. Second, it is recognized that the sample size necessary to detect differences in ordinal data as used in the SART and NASA TLX surveys would need to be much larger to detect differences. However, the ability and capacity to facilitate a sample size larger than 15 participants were not possible at the time of the study.

\section{Results}

    \subsection{Analysis for Number of Failures}
    
    To determine the cause of failures, we used an Exhaustive CHAID (Chi-Square Automatic Interaction Detection) test \cite{ExChi}. The dependent variable was the number of failures, and three explanatory variables were used in the algorithm: assembly type (antenna module or USB module), assembly location (local or remote) and assembly unit (unit 1, 2, or 3). Results of this test are displayed in Figure 8. 
    \begin{figure}[]
        \begin{center}
        \includegraphics[width=0.435\textwidth]{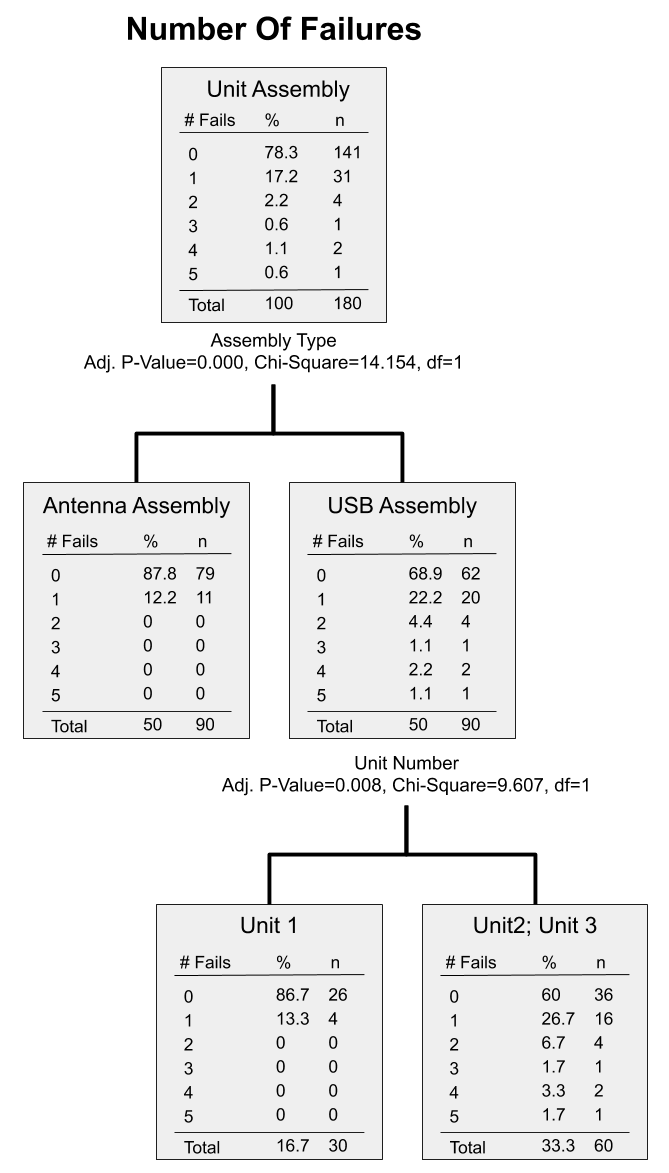}
        \caption{This decision tree depicts the percentage of failures between unit types as well as unit numbers for the USB module only.}
        \end{center}
        \label{fig:NumberFails}
    \end{figure}
    
    The first split with a p-value of less than 0.001 indicated that the strongest predictor of failure was assembly type. Operators were more likely to encounter a failure when deploying the USB module than when deploying an antenna module. The second strongest predictor of failure with a p-value of 0.008 was the assembly unit within the USB assembly type. Operators were more likely to encounter a failure when deploying the second or third USB module than when deploying the first USB module. It is important to note that assembly location was not a statistically significant predictor of failure.

    \subsection{Analysis for Assembly Time}
    We logged time to completion for each individual antenna and USB module. Each logged assembly time has three conditions, assembly type, assembly unit, and assembly location. We used a repeated measures ANOVA on the mean levels of assembly time, including the factors of assembly unit, assembly type and assembly location to test several hypotheses \cite{anova}. The null hypotheses stated that the mean of assembly time between the factors of assembly type, assembly location and assembly units were equivalent, and no interactions between factors exist. The alternative hypotheses stated that the mean of assembly time between the factors of assembly type, assembly location and assembly units were not equivalent. The factor of participant was blocked so the effects of its variability were not included in the error term. Based on the results of the ANOVA on the mean, several of the null hypotheses were rejected at the 95\% confidence level including assembly type F(1, 154) = 23.333, p = 0.000003, assembly unit F(2, 154) = 4.5914, p = 0.011568, assembly location F(2, 154) = 8.838, p = 0.003425 and the alternative hypothesis was accepted.
        
\begin{figure}[]
        \centering
        \includegraphics[width=0.45\textwidth]{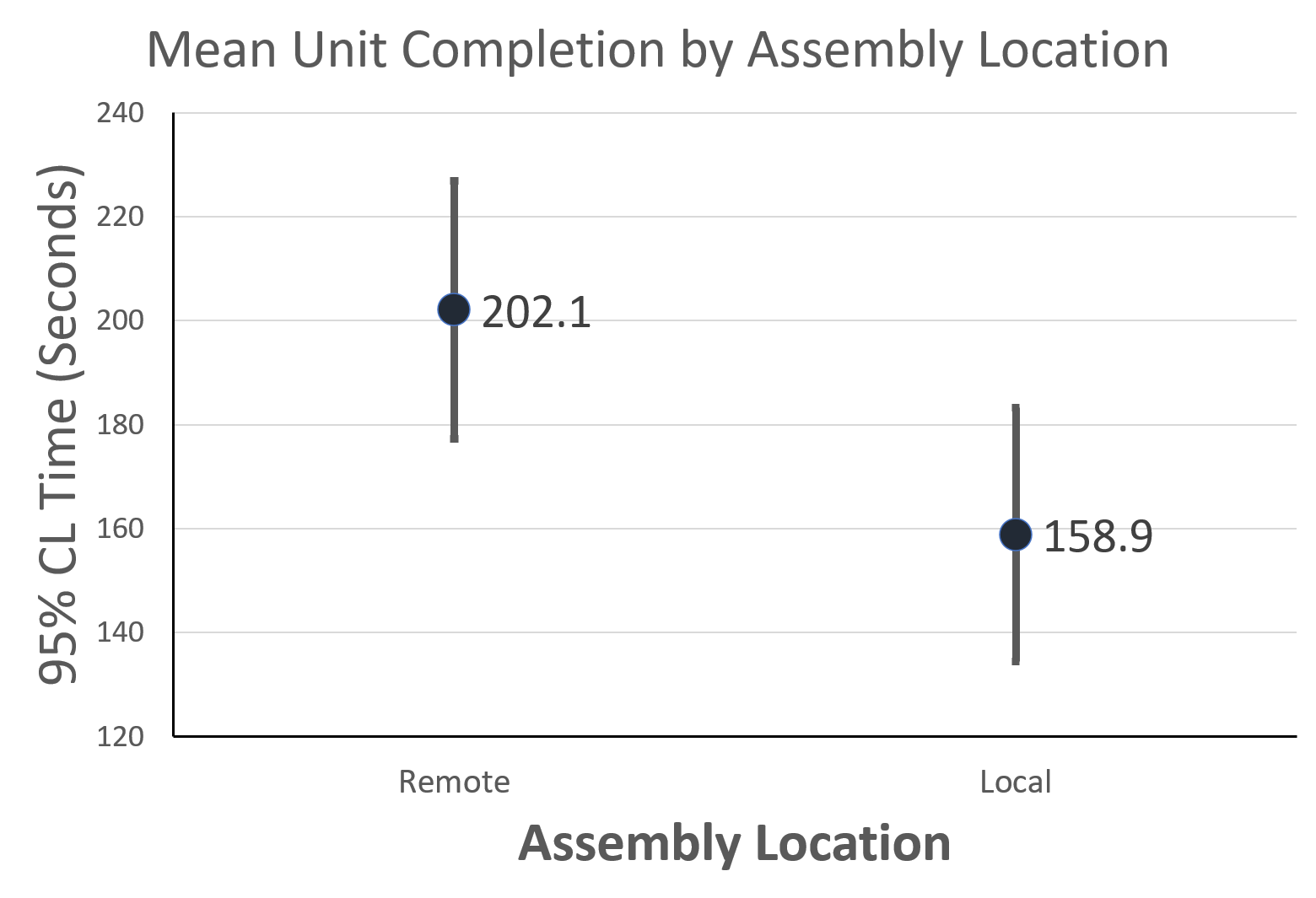}
        \caption{Means Plot of Time to completion by Assembly Location}
        \label{fig:assemloc}
    \end{figure}

    There was a statistically significant difference for the treatment of assembly location with respect to mean levels of assembly time. The mean assembly time in the remote setting was statically longer than the mean assembly time in the local setting (Figure \ref{fig:assemloc}).
        
\begin{figure}[]
        \centering
        \includegraphics[width=0.45\textwidth]{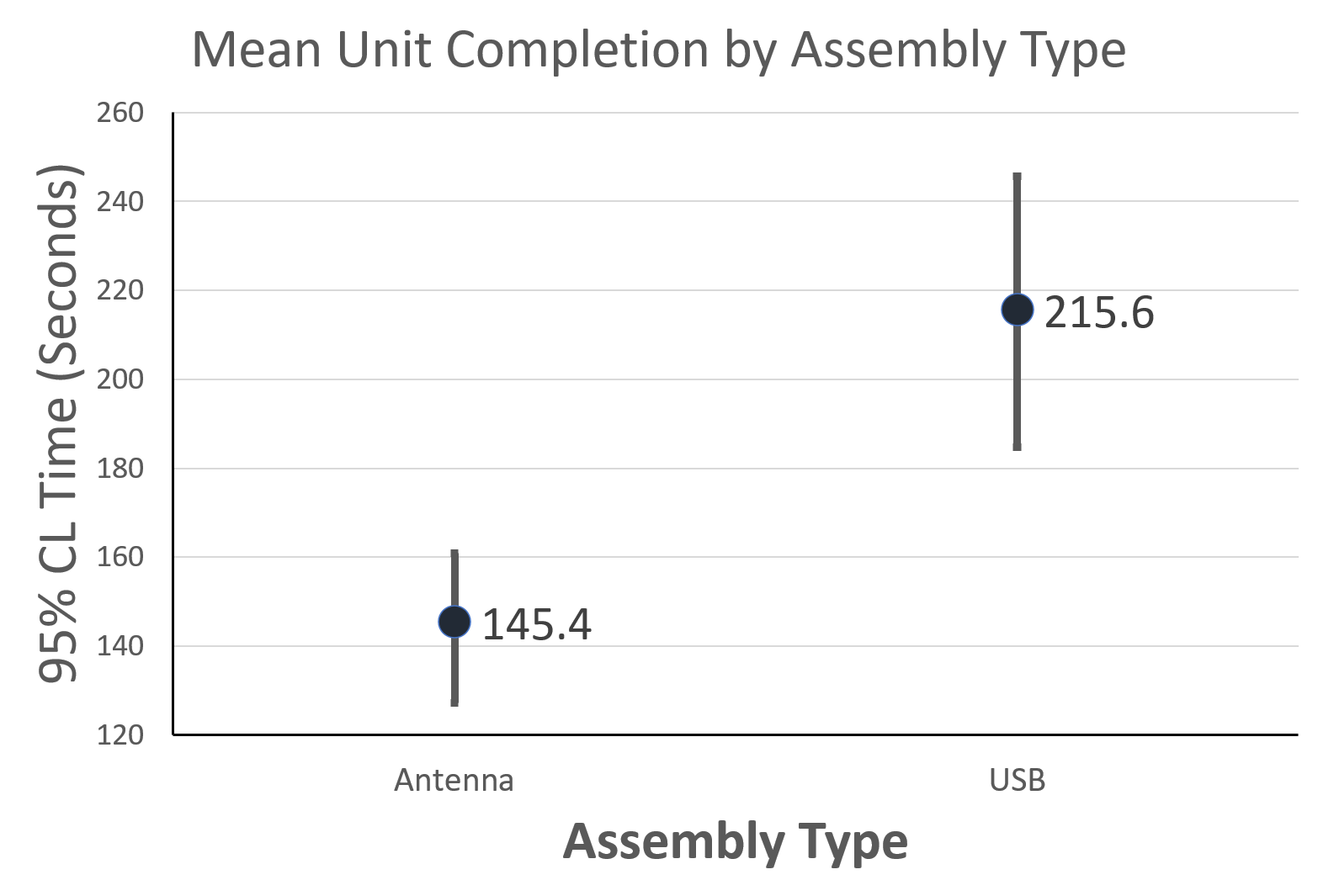}
        \caption{Means Plot of Time to Completion by Assembly Type}
        \label{fig:ModTimeComp}
    \end{figure}

    In addition to assembly location, the true means for assembly time by assembly type were also significantly different. The antenna assembly took less time to assemble than the USB assembly. Figure \ref{fig:ModTimeComp} shows a means plot of the true means of assembly time by assembly type.
        
        \subsubsection{Post-Hoc Analysis}
        
\begin{figure}[H]
        \centering
        \includegraphics[width=0.45\textwidth]{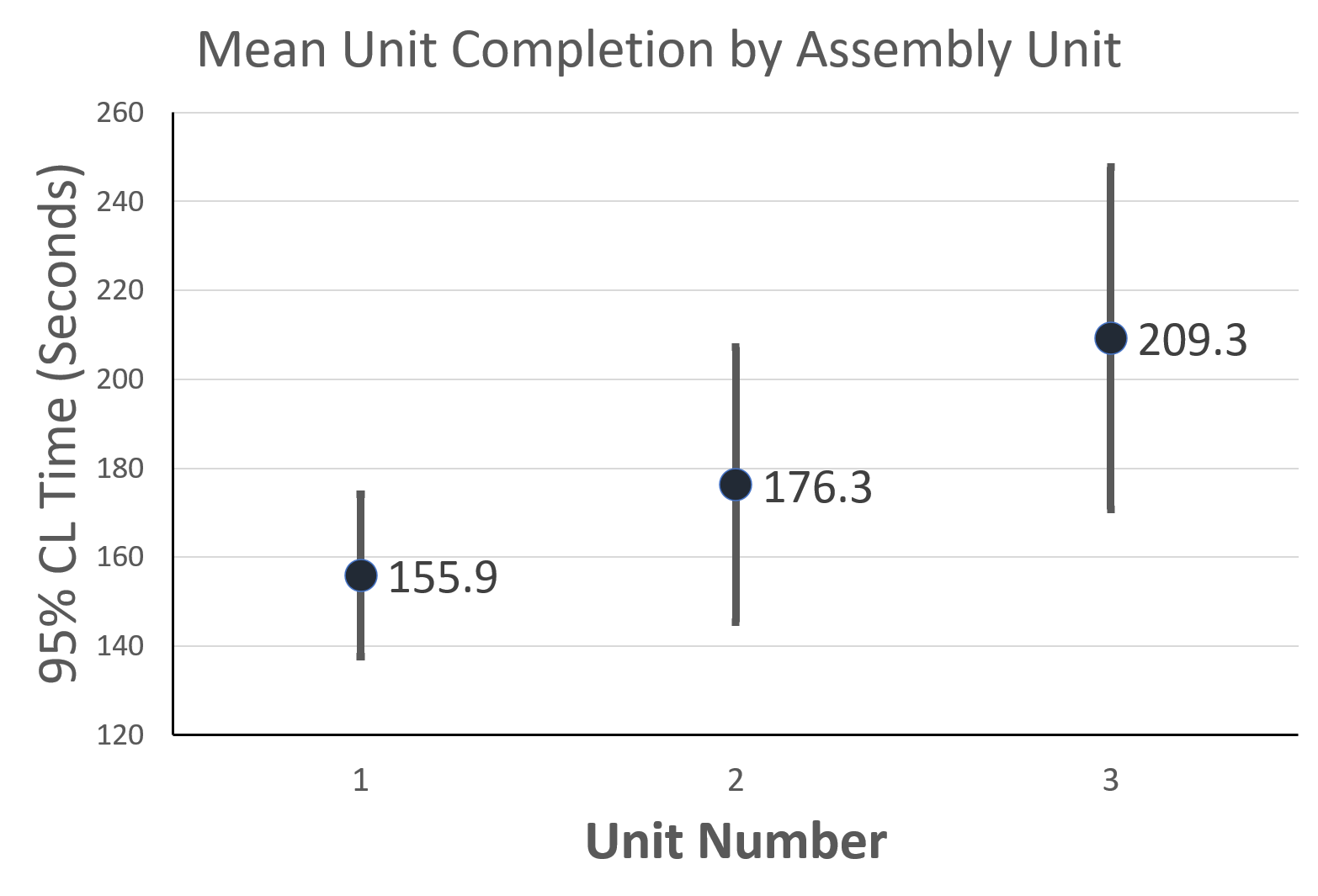}
        \caption{Means Plot of Time to completion by Assembly Unit Number}
        \label{fig:UnitTimeComp}
    \end{figure}

        Since the mean times to completion were not equivalent between assembly units, we conducted a post-hoc analysis to determine which groups were different from each other. We evaluated all pairwise comparisons using the Games-Howell post-hoc test on the means due to unequal variance in the factor of assembly unit \cite{Pairwise}. Table \ref{tab:UnitMean} shows the results of the Games-Howell post-hoc test on the sample means between all assembly units. The sample mean for Units 1 and 2 were equivalent at the 95\% confidence level.  Units 2 and 3 were equivalent at the 95\% confidence level. The sample mean of Unit 1 was significantly different than Unit 3 at the 95\% confidence level. Figure \ref{fig:UnitTimeComp} shows the means plot of the true means at the 95\% confidence level of the mean assembly time for each of the three units.
        
        \begin{table}[H]
            \centering
            \caption{Assembly Unit: Post-Hoc Analysis}
            \fboxsep=0mm
            \fboxrule=0.4pt
            \fbox{\includegraphics[width=0.45\textwidth]{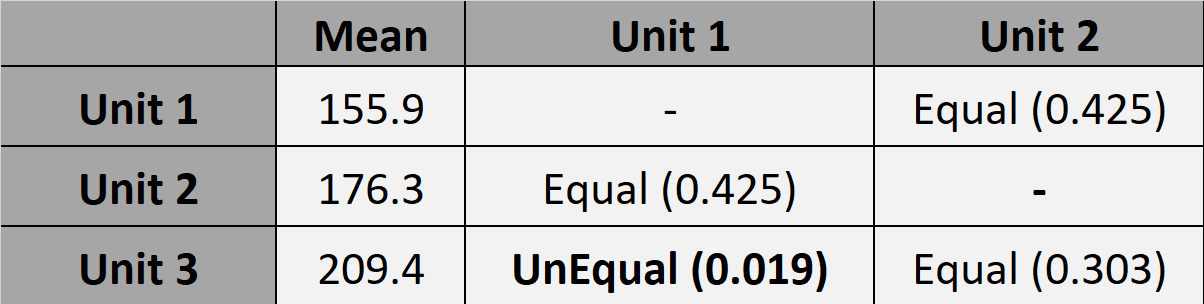}}
            \label{tab:UnitMean}
        \end{table}
    
    \subsection{Analysis of Subjective Measurements}
    
    Survey results were used to compare the SA and CL of the operator between the local and remote assembly locations. We analyzed the cumulative scores of the surveys as well as the responses to individual question.
    
    \newpage
    \vspace{500mm}
        
        \subsubsection{Situational Awareness Rating Technique}
        
        After using an Anderson Darling Test \cite{ADtest} to confirm that all SART data was drawn from the same sample population, we applied a Kruskal-Wallis ANOVA \cite{kwanova} to determine variation of situation awareness between the remote and local environment. In addressing the cumulative scores, no evidence of a significant variation was found between the remote and local SART scores, with a resultant p-value of 0.2432. Since this disagreed with the performance metrics, we analyzed survey responses to individual questions in attempt to find the source of the disagreement. After performing a Kruskal-Wallis ANOVA on each question, one question showed greater SA in the local environment at the 95\% confidence level. However, four more questions showed greater SA in the local environment at the 85\% confidence level. 
        
        \subsubsection{NASA Task Load Index}
        
        When analyzing the NASA TLX, we used a similar procedure to the SART analysis. We first used an Anderson Darling Test \cite{ADtest} to ensure that all of the collected data was drawn from the same sample population. We then utilized a Kruskal-Wallis ANOVA \cite{kwanova} in search of variation between the remote and local environments. Both cumulative and individual question scores showed no significant variation.


\section{Discussion}

    The primary goal of this experiment was to develop a methodology to accurately assess the human factors associated with teleoperated assembly tasks. We believed that our methodology was successful if the results showed a positive correlation between subjective measurements of SA and performance metrics and a negative correlation between subjective measurements of CL and performance metrics. In addition to the correlations, we were also looking for an increase in SA from the remote to local environment. The performance metrics showed an increase in SA and a decrease in CL from the remote to local environment, but we did not see the expected correlations between the performance metrics and the subjective measurements of SA and CL. Results from the performance metrics show that the assembly task was sufficiently difficult to highlight changes in SA and CL between environments. However, the results from the subjective measurements of SA and CL revealed the need for changes in the methodology moving forward.

    \subsection{Surveys}
    Results from the SART and NASA TLX did not meet our criteria for a successful methodology. A successful methodology would have shown a consistent increase in SA from the remote to local setting. While we did observe some statistically significant increase in SA between settings, it was not enough to convince us that the methodology was accurately assessing human factors. The SART and TLX use a procedure that takes answers from individual questions, and equates them to a final cumulative score. Our original method of analysis only accounted for each participant's final score. This yielded results stating that there was no statistical difference in SA and CL between the local and remote environments. 
    
    We then conducted further investigation of each survey, to find an explanation for these results. Further examination of the SART and TLX surveys revealed that some questions could have multiple interpretations and other questions contained wording that could confuse the participant. For example, question nine of the SART asks, ``How useful is the received information with respect to achieving your task?'' It is our belief that participants may have had different understandings of what was considered as ``received information.'' In the local setting, some participants may have only considered information on the user interface as ``received information'' while other participants may have considered their eyesight of the operating room and the information on the user interface as ``received information'' which is how we intended participants to interpret the question. Since the time to completion in the local environment was faster than the remote environment, results from this SART question should have shown a statistical increase in information utility from the remote to local environment as well. When we designed this experiment, we planned to analyze the cumulative scores from the surveys in order to quantify the subjective measurements of SA and CL, so we did not alter any of the surveys. However, certain questions from the official surveys contained confusing language and did not feel applicable such as the question gauging ``arousal'' in the situation from the SART survey. This made us believe that the cumulative scores from the surveys were normalized, and that we should proceed with a different method of analysis.
    
        \subsubsection{SART} 
        We decided to analyze the SART on a question-by-question basis because we believed the cumulative scores were normalized. This analysis revealed trends of greater SA in the local environment, but it was not enough to convince us that the methodology was accurately assessing SA. We believe that editing the wording of questions in the survey and adding more participants to future experiments will make the subjective measurements of SA align with the performance metrics.
        
        \subsubsection{NASA TLX}
        Following a similar approach to our SART analysis, we decided to analyze each dimension of the TLX individually, rather than analyzing the adjusted ratings. The six scales assessed are mental demands, physical demands, temporal demands, performance, effort, and frustration. Previous studies have identified the benefits of analyzing the individual dimensions of the NASA TLX to better assess the factors that contribute to workload \cite{TLXDimensions}. Each question was analyzed between remote and local environments. Histograms of the responses coupled with mean response values revealed trends of a greater CL in the remote environment, but this was not supported by the statistical analysis with no significant results. A final factor to consider is the workload and fatigue put into filling out the surveys. According to Noyes and Bruneau, a computer based administration of the NASA TLX causes a greater workload on the subject than a paper-based survey \cite{TLXDimensions}. This could potentially lead participants to rush through the survey, invalidating the results. We believe adding more participants and physically administering the survey will make the subjective measurements of CL correlate with the performance metrics as expected.
        
        \subsubsection{SUS}
        Results of the SUS help us determine what parts of our system should be improved for more effective teleoperation. We calculated an average score of 73.8 from the participants' responses which ranks the TSS slightly above average \cite{sus}. This result tells us that our hardware and software are adequate to execute this methodology. However, there is still room for system improvements in future experiments.

    \subsection{Applications to Lunar Missions}
    Though this pilot study was intended to determine an effective methodology for assessing human factors associated with telerobotic assembly tasks, the experiment still provided useful insights that could be applied to future lunar missions. For example, our results showed that there is a similar success rate between local and remote operation. However, the task is completed significantly faster when operating locally. This further supports the claim that lunar teleoperated assembly tasks are viable from the Gateway (remote), albeit less efficient than if performed from the lunar surface (local). To increase the efficiency of teleoperation, a third person camera could be placed in a location viewing the rover and the assembly task. This camera is analogous to the participant's vision of the operating room in the local phase of our experiment. We believe that direct observation of the operating room strongly contributed to the decreased time to completion from the remote to local environment. Quick assembly will be imperative to future lunar missions, as assembly during the lunar night would be very difficult due to lighting and rover operational constraints.
    
    \subsection{Future Work}

    Moving forward, important procedural changes to our experiment have become evident. To improve the methodology, we need to adjust the subjective assessment of SA and CL to ensure it is representative of the operator's experience and it aligns with the performance metrics. In the next iteration of this experiment, we will alter the NASA TLX and SART surveys to ensure every question is interpreted in the same way and relevant to the experiment. We will also consider adding new survey(s) to assess human factors. One survey in specific is the Situational Awareness for SHAPE (SASHA) which not only assesses SA but also workload and trust in the system \cite{SASHA}. Since the SART primarily assess SA and the TLX primarily assess workload, we could verify results from the SART and TLX by comparing them to results from SASHA. To ensure the surveys accurately represent the operator's experience, we will interview participants after they have completed the experiment regarding their thoughts on the surveys. During this interview, we will ask the participants to describe the change (if any) in SA and CL between the remote and local environment.

    In addition to subjective measurement changes, we will revamp the training phase of our experiment to ensure that each participant is receiving the exact same training prior to the experiment. We will do this by creating a training video, which will be watched by each participant prior to the experiment. In our current methodology there was not a strict script for the training phase, but rather identified topics that needed to be explained to the participant. The word-by-word explanation for these topics could have varied between participants considering the training phase was not always delivered to the participant by the same researcher.

    After ensuring our methodology is accurately assessing the human factors associated with telerobotic assembly tasks, new variables can be introduced to the experiment. Some variables to consider are varied latency, varied frame rate, and new camera perspectives. Previous research has found a threshold frame rate of five frames per second while teleoperating a rover for geological exploration \cite{Ben}. We would like to apply this to an assembly task, which requires more precision than a surveying task.

    In terms of new perspectives for teleoperation, virtual reality is a promising option. By adding a stereoscopic camera to a rover, the user would have the ability to see as if looking directly though the eyes of the rover. An element of augmented reality could also come into play, turning the user interface into a more informative head-up display. This would allow operators to see everything in three dimensions, without looking between two 2-D video feeds. Perspective can also be added by allowing the users a camera view providing a third person view of the assembly room and rover, to mimic the point of view seen when locally operating a robot. Perfecting the methodology described in this paper would allow us assess new and innovative methods of teleoperation for assembly tasks which will significantly aid the colonization of the Moon.    

\section{Conclusion}

We developed a methodology with the goal of assessing the human factors associated with teleoperated assembly tasks. Participants used our rover control system, the TSS, to assemble a radio interferometer through local operation of the rover and remote teleoperation of the rover. We measured performance metrics including time to completion, number of failures, and unit placement and administered surveys to assess the operators SA and CL in both environments. To validate our subjective assessment of SA and CL, we looked for correlations between the subjective measurements and the performance metrics. The performance metrics suggested greater SA and lower CL in the local environment compared the remote environment. However, the subjective measurements did not provide sufficient evidence of variation of SA and CL between the two environments. We believe the subjective measurements did not accurately assess the operator's SA and CL. Because of the disagreement between our measurements, we are modifying our methodology for future experiments. By reassessing each survey, we can make sure that all questions are posed in such a way that will not create any confusion. We will also add more participants to the experiment in order to amplify any variation in SA and CL between environments. Once an accurate methodology has been developed, we will be able to assess the effects of new variables on an operator's SA and CL while performing telerobotic assembly.


\acknowledgments
This work is directly supported by the NASA Solar System Exploration Virtual Institute cooperative agreement 80ARC017M0006. We would also like to acknowledge our non-author contributors, Dan Szafir, Michael Walker, Midhun Menon, and Joseph Minafra, for their assistance with our research.


\bibliography{IEEEabrv,references}
\bibliographystyle{IEEEtran}

\thebiography

\begin{biographywithpic}
{Arun Kumar}{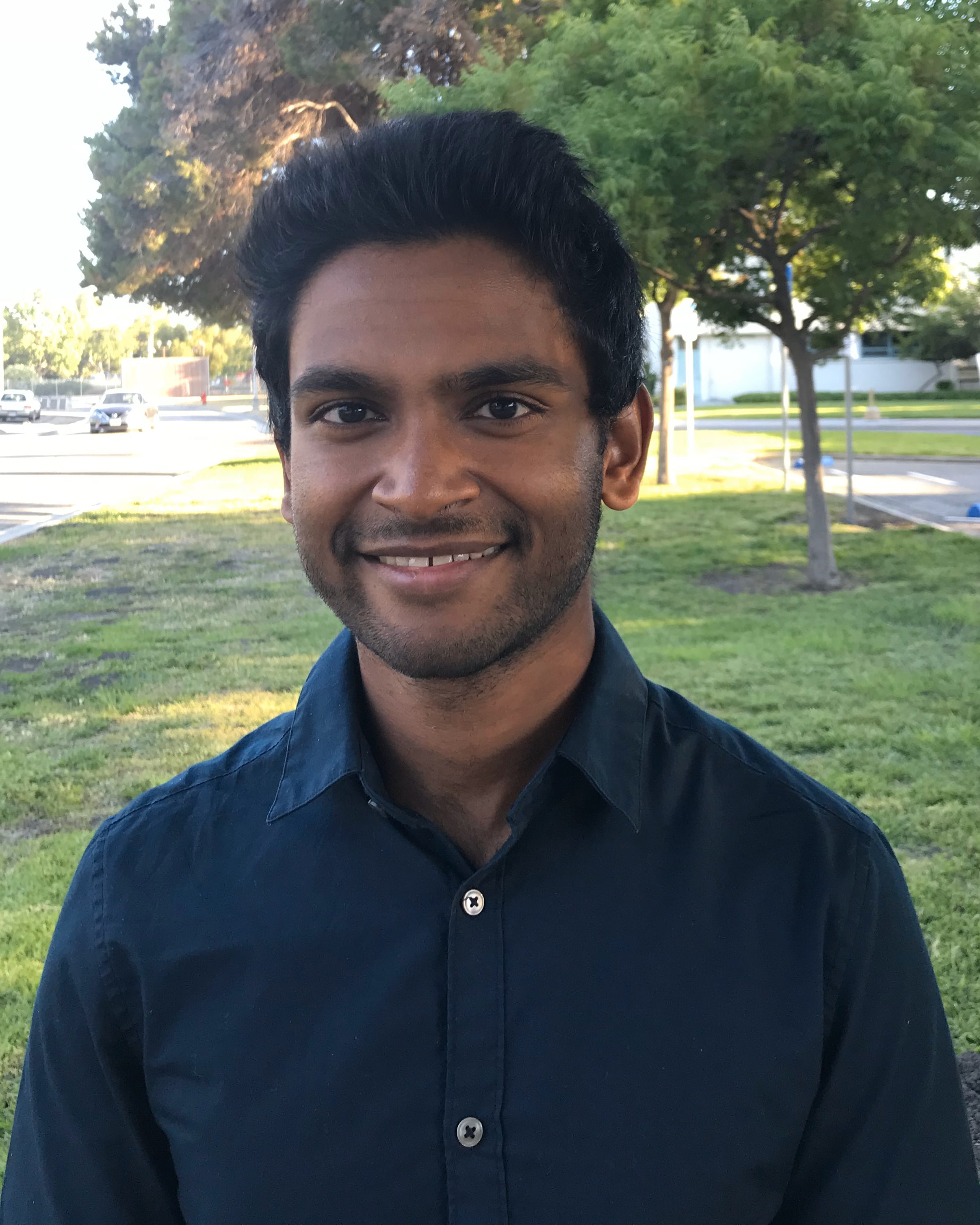}
is pursuing a B.S. degree in Mechanical Engineering at the University of Colorado Boulder and is expected to graduate in May 2020. He is the manager of the Telerobotics Lab in the Center for Astrophysics and Space Astronomy. He is also a member of the NASA SSERVI Network for Exploration and Space Science (NESS) team.
\end{biographywithpic} 

\begin{biographywithpic}
{Mason Bell}{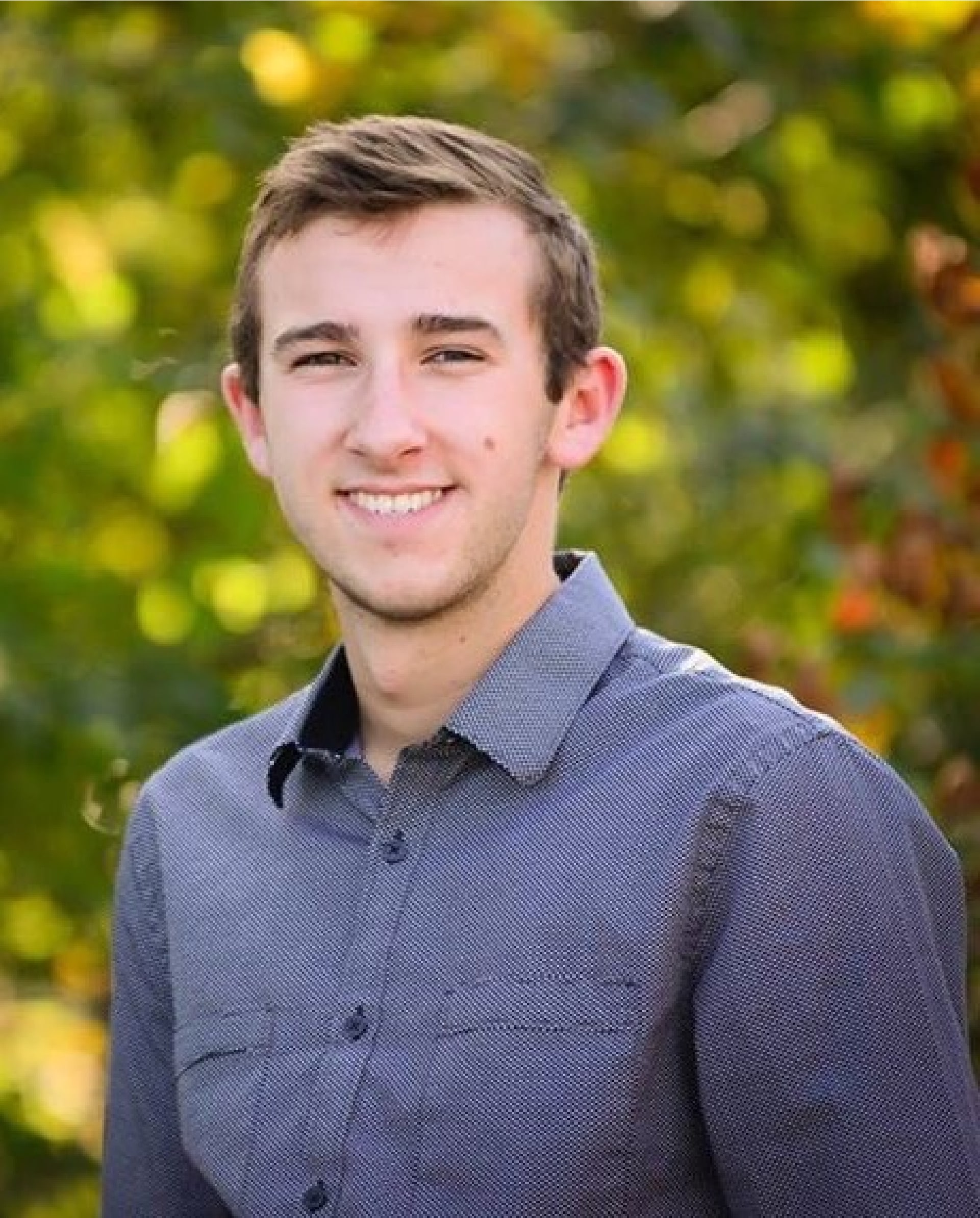}
is currently pursuing a B.S. degree in Computer and Electrical Engineering at the University of Colorado Boulder, and is expected to graduate in May 2021. He is a member of the Center for Astrophysics and Space Astronomy’s undergraduate telerobotics group. He is also a member of the NASA SSERVI Network for Exploration and Space Science (NESS) team.
\end{biographywithpic} 

\begin{biographywithpic}
{Benjamin Mellinkoff}{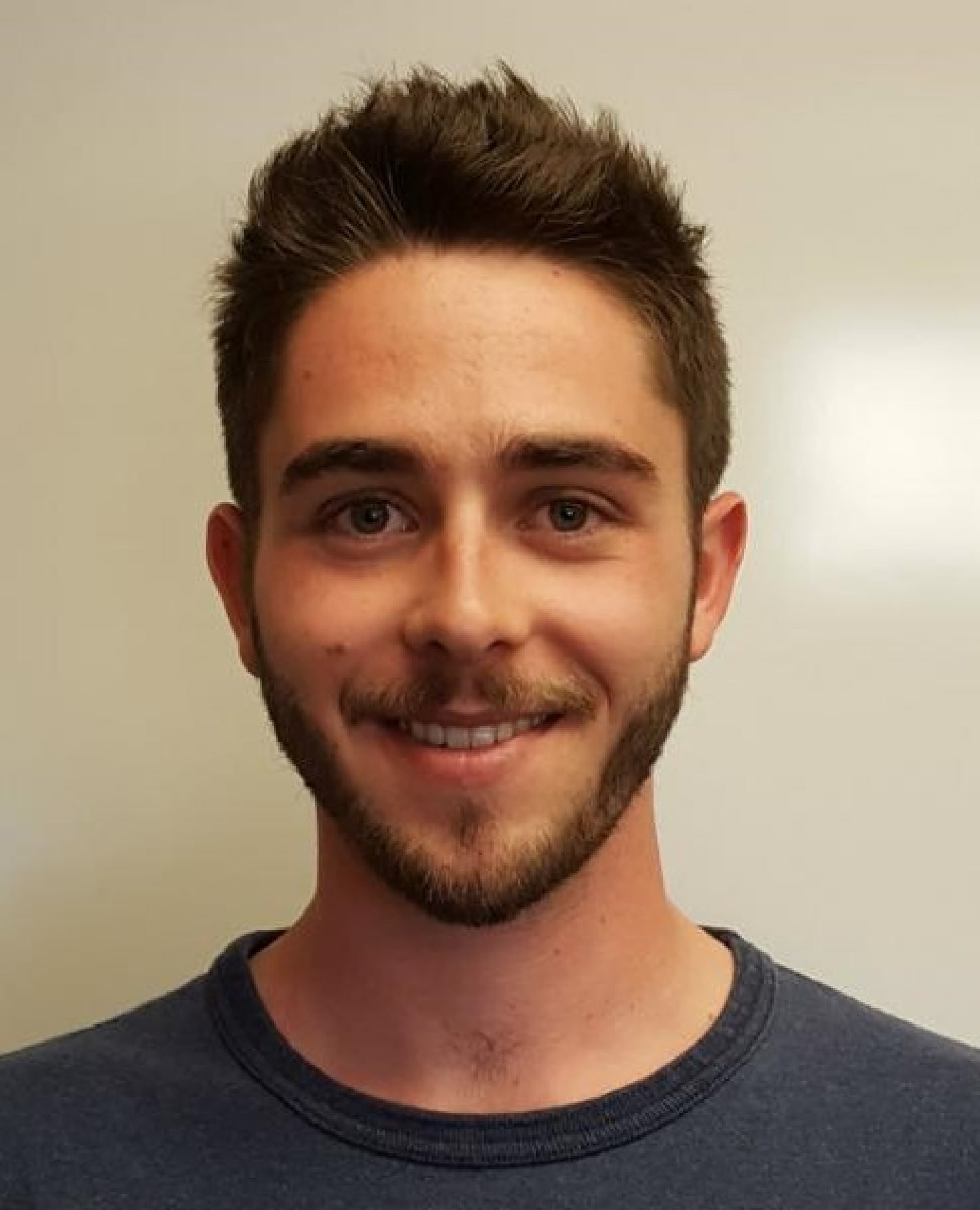}
graduated from the University of Colorado Boulder with a B.S. and M.S. degree in Aerospace Engineering with an emphasis in Aerosp-ace Systems: Controls. Before graduation, he worked as the manager of the Center for Astrophysics and Space Astronomy's Telerobotics Lab. He is also a member of the NASA SSERVI Network for Exploration and Space Science (NESS) team.
\end{biographywithpic} 

\begin{biographywithpic}
{Alex Sandoval}{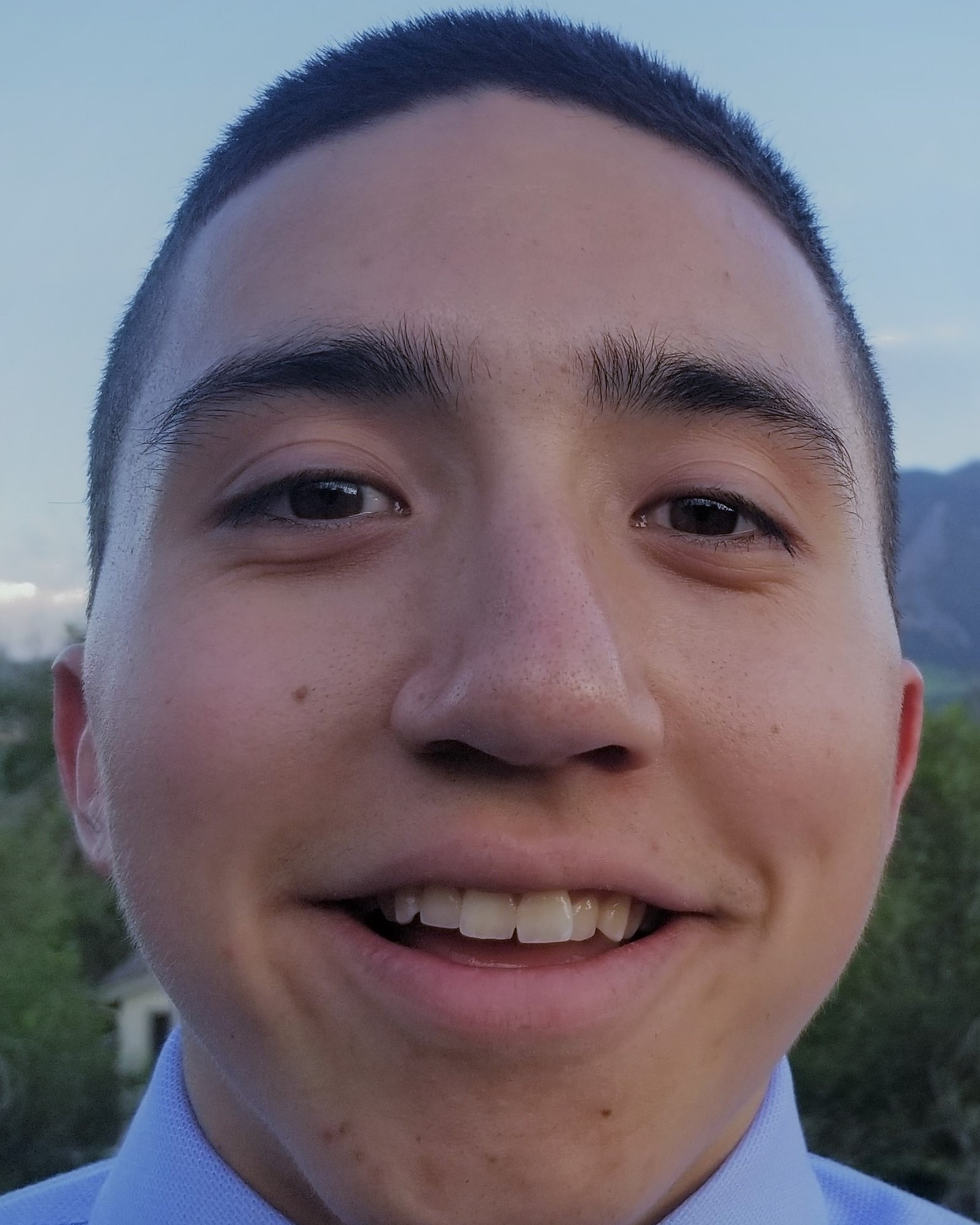}
received his B.S. in Electrical and Computer Engineering from the University of
Colorado Boulder in 2019. His interest in aerospace started after taking notice of the growing environment among space companies, space activities, and space people making an impact across the industry. After supporting a space robotics research group along with other rover, radio, and embedded work, he has started a role at Ball Aerospace as a robotics software engineer for a research and development program trying to start to make an impact as well. 
\end{biographywithpic} 

\begin{biographywithpic}
{Wendy Bailey Martin}{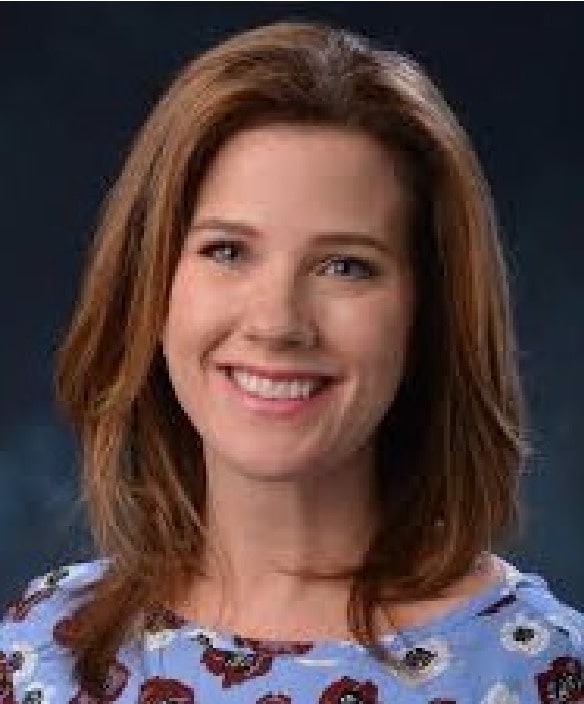}
is the W. Edwards Deming Professor of Management in the Lockheed-Martin Engineering Management Program, teaching in the area of Quality Science. She earned her undergraduate degree in Mechanical Engineering from Purdue University, and a Masters of Engineering from the Lockheed-Martin Engineering Management Program at the University of Colorado Boulder, with an emphasis in six sigma, quality systems and applied statistics. Prior to her graduate degree, she was trained in statistical methods by Luftig Warren International (LWI). Wendy also worked for 14 years at Anheuser-Busch, where she became skilled in the application of statistics in an industrial environment.
\end{biographywithpic}

\begin{biographywithpic}
{Jack Burns}{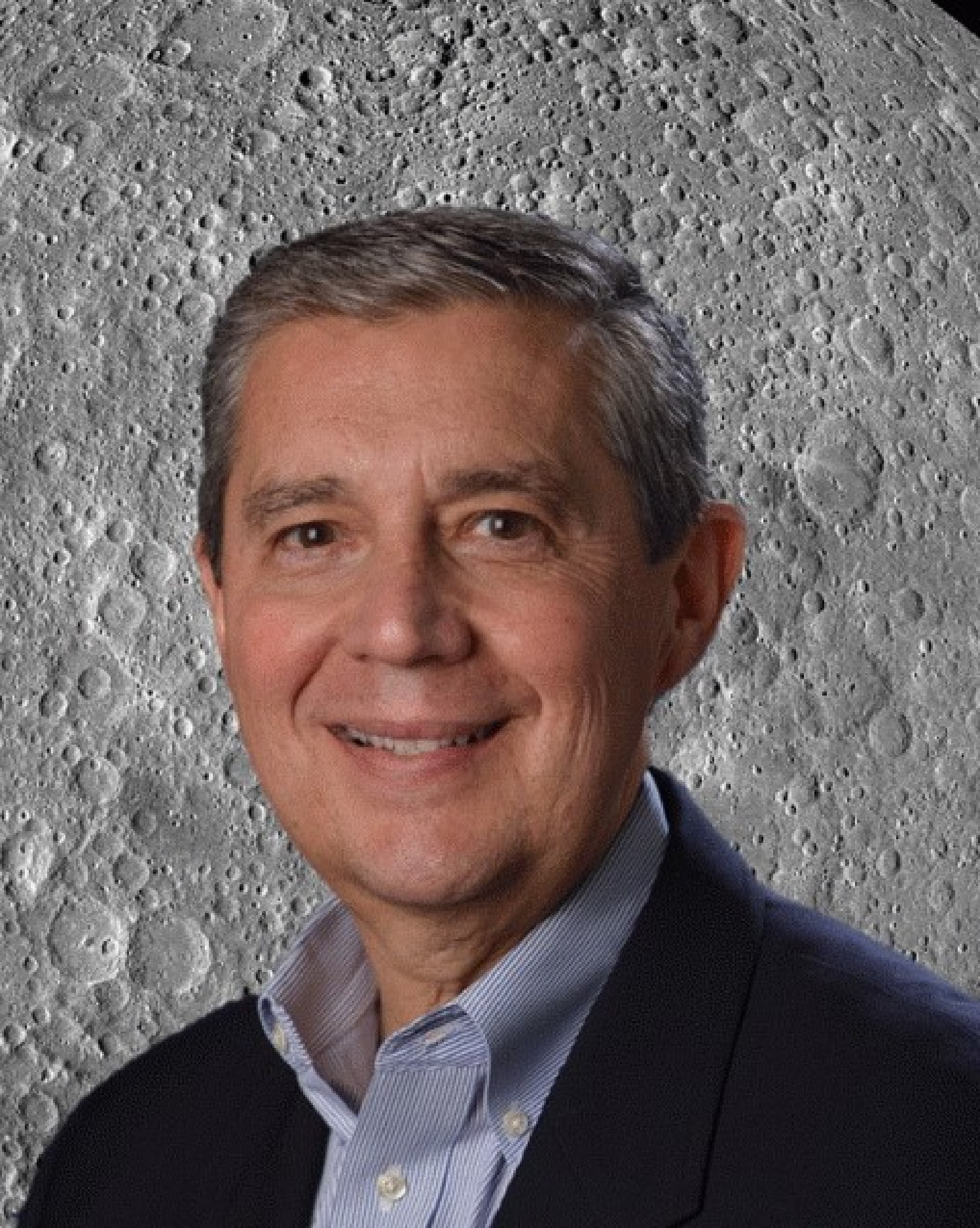}
is a Professor in the Department of Astrophysical and Planetary Sciences and Vice President Emeritus for the University of Colorado. He is also Director of the NASA-funded SSERVI Network for Exploration and Space Science (NESS). Burns is an elected Fellow of the American Physical Society and the American Association for the Advancement of Science. He received NASA’s Exceptional Public Service Medal in 2010 and NASA’s Group Achievement Award for Surface Telerobotics in 2014. Burns was a member of the Presidential Transition Team for NASA in 2016/17. Burns recently served as senior
Vice President of the American Astronomical Society.
\end{biographywithpic} 

\end{document}